\documentclass[conference]{IEEEtran}
\IEEEoverridecommandlockouts
\usepackage{cite}
\usepackage{amsmath,amssymb,amsfonts}
\usepackage{algorithmic}
\usepackage{graphicx}
\usepackage{textcomp}
\usepackage{csquotes}
\usepackage{xcolor}
\usepackage{adjustbox}
\usepackage{subcaption}
\usepackage{hyperref}
\usepackage{multirow}
\usepackage{orcidlink}

\def\BibTeX{{\rm B\kern-.05em{\sc i\kern-.025em b}\kern-.08em
    T\kern-.1667em\lower.7ex\hbox{E}\kern-.125emX}}

\begin{document}

\title{Enhancing Underwater Navigation through Cross-Correlation-Aware Deep INS/DVL Fusion}

\author{\IEEEauthorblockN{Nadav Cohen\IEEEauthorrefmark{1}\orcidlink{0000-0002-8249-0239} and Itzik Klein \orcidlink{0000-0001-7846-0654}}
\IEEEauthorblockA{\textit{The Hatter Department of Marine Technologies,} \\
\textit{Charney School of Marine Sciences,}\\ \textit{University of Haifa}\\
\textit{Haifa, Israel}}

\thanks{\IEEEauthorrefmark{1}Corresponding author: N. Cohen (email: ncohe140@campus.haifa.ac.il).}}

\maketitle

\begin{abstract}
The accurate navigation of autonomous underwater vehicles critically depends on the precision of Doppler velocity log (DVL) velocity measurements. Recent advancements in deep learning have demonstrated significant potential in improving DVL outputs by leveraging spatiotemporal dependencies across multiple sensor modalities. However, integrating these estimates into model-based filters, such as the extended Kalman filter, introduces statistical inconsistencies, most notably, cross-correlations between process and measurement noise. This paper addresses this challenge by proposing a cross-correlation-aware deep INS/DVL fusion framework. Building upon BeamsNet, a convolutional neural network designed to estimate AUV velocity using DVL and inertial data, we integrate its output into a navigation filter that explicitly accounts for the cross-correlation induced between the noise sources. This approach improves filter consistency and better reflects the underlying sensor error structure. Evaluated on two real-world underwater trajectories, the proposed method outperforms both least squares and cross-correlation-neglecting approaches in terms of state uncertainty. Notably, improvements exceed 10\% in velocity and misalignment angle confidence metrics. Beyond demonstrating empirical performance, this framework provides a theoretically principled mechanism for embedding deep learning outputs within stochastic filters.
\end{abstract}


\section{Introduction}
\noindent
Autonomous underwater vehicles (AUVs) are valuable tools for exploring and operating underwater. They are used in a wide range of applications, including seafloor mapping, underwater construction and inspection, environmental monitoring, and studying marine life \cite{nicholson2008present,griffiths2002technology}.  To accomplish their task accurately and robustly, precise navigation is required. Commonly, this is achieved by fusing inertial sensors with a Doppler velocity log (DVL) \cite{miller2010autonomous}. The DVL is an acoustic sensor that utilizes the Doppler effect. This device transmits four acoustic beams to the seafloor, which are then reflected back. Based on the frequency shift, the DVL calculates the velocity of each beam and then estimates the velocity vector of the AUV \cite{rudolph2012doppler}.
\\ \noindent
Data-driven methods have been employed in AUV-related tasks with promising outcomes \cite{cohen2024inertial, zhang2025underwater, 9799796, topini2023experimental, makam2024spectrally}. DCNet, a data-driven framework that utilizes a two-dimensional convolution kernel in an innovative way, demonstrated its ability to improve the process of DVL calibration~\cite{ YAMPOLSKY2025104525}. Deep-learning frameworks have been used to estimate real-world scenarios of missing DVL beams in partial and complete outage scenarios~\cite{ yona2024missbeamnet, cohen2023set}. Additionally, when all beams are available, the BeamsNet approach~\cite{ cohen2022beamsnet} offers a more accurate and robust velocity solution using a dedicated deep-learning framework. Deep learning methods were suggested for the fusion process to adaptively estimate the process noise covariance in the inertial DVL fusion process~\cite{ cohen2025adaptive, levy2025adaptive }. Recently, an end-to-end deep learning approach was suggested to estimate the AUV acceleration vector, which is introduced to the navigation filter as an additional measurement~\cite{ YS2025}. 
In normal operating conditions of the DVL, the BeamsNet approach outperforms model-based approaches. This method employs both inertial measurements and past DVL measurements to estimate the current velocity vector. When updating the navigation filter with this measurement, a cross-correlation arises between the BeamsNet velocity vector measurement and the inertial-based process noise. This process-measurement cross-correlation matrix should be taken into account in the navigation filter to obtain a desired matched filter~\cite{ simon2006optimal}.  
\\ \noindent
In this paper, we employ the inertial/DVL navigation filter based on the extended Kalman filter (EKF), taking into account the process-measurement cross-correlation matrix. The latter is calculated using the inertial and BeamsNet error sources. Using two real-world underwater AUV datasets, we show the necessity of the cross-correlation matrix to allow filter consistency and robustness. 
\\ \noindent
The rest of the paper is organized as follows: Section \ref{PF} formulates the problem and presents the theoretical foundation for incorporating cross-correlations within the extended Kalman filter. Section \ref{CS} introduces the proposed cross-correlation-aware deep INS/DVL fusion framework, detailing the integration of BeamsNet with a modified filter formulation. Section \ref{res} presents the experimental results and performance analysis based on real-world AUV trajectories. Finally, Section \ref{con} concludes the findings in the paper.

\section{Problem Formulation}\label{PF}
\subsection{EKF with Correlated Noise}\label{sec:EKFcorr}
\noindent
In the classical derivation of the error-state EKF, it is typically assumed that the process noise and measurement noise are uncorrelated. This is due to the fact that the different sensors provide the information for the process and update. However, in some cases, like the one we introduce in this paper, non-negligible cross-correlations may exist between the process and measurement noise terms. In this section, we present an error-state EKF framework that accounts for such correlation, following the modified Kalman filter equations that explicitly incorporate this dependency.
\\ \noindent
Let the system dynamics and measurement model be described by the discrete-time equations \cite{bar2004estimation,groves2013principles}:
\begin{equation}
\boldsymbol{x}_k = \mathbf{F}_{k-1} \boldsymbol{x}_{k-1} + \mathbf{G}_{k-1} \boldsymbol{w}_{k-1},
\end{equation}
where \( \boldsymbol{x}_k \) denotes the system state vector at time step \( k \), \( \mathbf{F}_{k-1} \) is the state transition matrix that propagates the state from time \( k-1 \) to \( k \), and \( \mathbf{G}_{k-1} \) is the process noise input matrix.\\
\noindent
The measurement model is:
\begin{equation}
\boldsymbol{y}_k = \mathbf{H}_k \boldsymbol{x}_k + \boldsymbol{v}_k,
\end{equation}
\noindent
where, \( \boldsymbol{y}_k \) is the measurement vector at time step \( k \) and \( \mathbf{H}_k \) is the observation matrix that maps the state vector to the measurement space. Additionally, \( \boldsymbol{w}_k \sim \mathcal{N}(\boldsymbol{0}, \mathbf{Q}_k) \) denotes the process noise with associated process noise covariance $\mathbf{Q}_k$, and \( \boldsymbol{v}_k \sim \mathcal{N}(\boldsymbol{0}, \mathbf{R}_k) \) denotes the measurement noise with associated measurement noise covariance $\mathbf{R}_k$ . \\
\noindent
The cross-correlation matrix between the process and measurement noise covariances is defined as~\cite{simon2006optimal}:
\begin{equation}\label{eqn:M}
\mathbb{E}[\boldsymbol{w}_k \boldsymbol{v}_j^\top] = \mathbf{M}_k \delta_{k-j+1},
\end{equation}
\noindent
indicating that the process noise at time \( k \) is correlated with the measurement noise at time \( k + 1 \). This structure arises naturally in systems where the same external disturbance influences both the system dynamics and the measurement process, albeit with a one-step time lag.
\\ \noindent
To incorporate this cross-correlation into the Kalman gain computation, the innovation covariance must be adjusted such that the Kalman gain becomes:
\begin{align}\label{eqn:KalmanGain}
\mathbf{K}_k = {} & \left( \mathbf{P}_k^- \mathbf{H}_k^\top + \mathbf{M}_k \right) \nonumber \\
& \cdot \left( \mathbf{H}_k \mathbf{P}_k^- \mathbf{H}_k^\top + \mathbf{H}_k \mathbf{M}_k + \mathbf{M}_k^\top \mathbf{H}_k^\top + \mathbf{R}_k \right)^{-1},
\end{align}

\noindent
where \( \mathbf{P}_k^- \) is the prior error covariance matrix. 
\\
\noindent
Consequently, the posterior error covariance is updated according to:
\begin{equation}
\mathbf{P}_k = \mathbf{P}_k^- - \mathbf{K}_k \left( \mathbf{H}_k \mathbf{P}_k^- + \mathbf{M}_k^\top \right).
\end{equation}
These modified expressions account for the non-zero cross-correlation between process and measurement noise, improving filter consistency in scenarios where this assumption is violated. The rest of the error state EKF equations and process remain the same with the exception of the Kalman gain \eqref{eqn:KalmanGain} and can be seen in \cite{farrell2008aided}, for example.
\subsection{Cross-Correlation within INS/DVL Fusion}\label{sec:cross}
\noindent
Recent advances in deep learning have demonstrated significant potential in time series estimation and sensor fusion, especially in navigation systems where standard model-based filters often struggle with drift, nonlinearity, or degraded measurement conditions. By leveraging the expressive power of deep neural networks (DNNs), it is possible to model complex dependencies between sensor modalities and capture higher-order temporal patterns in the data. In particular, DNN-based approaches that jointly process inertial and acoustic measurements can yield more accurate velocity estimates than classical extended Kalman filtering alone.
\\ \noindent
Consider a system where the goal is to estimate the vehicle’s velocity using both the inertial sensors, which include accelerometers that provide the specific force vector \( \boldsymbol{f}_k \in \mathbb{R}^3 \) and gyroscopes that measure the angular velocity vector \( \boldsymbol{\omega}_k \in \mathbb{R}^3 \), and a DVL, which provides beam velocity measurements \( \boldsymbol{z}_k^{\mathrm{DVL}} \in \mathbb{R}^4 \). A deep neural network can be trained to produce a fused velocity estimate via a nonlinear function:
\begin{equation}
\boldsymbol{\hat{v}}_k = \mathcal{F}_\theta(\boldsymbol{f}_{k-T:k}, \boldsymbol{\omega}_{k-T:k}, \boldsymbol{z}^{\mathrm{DVL}}_{k-T:k}),
\end{equation}
where \( \mathcal{F}_\theta(\cdot) \) denotes the DNN with parameters \( \theta \) and the inputs consist of a temporal window of size \( T \). The output \( \boldsymbol{\hat{v}}_k \) is the estimated velocity vector at time \( k \), typically expressed in the body frame.
\\ \noindent
The challenge arises from the stochastic properties of the inputs. The IMU measurements \( \boldsymbol{f}_k \) and \( \boldsymbol{\omega}_k \) are driven by process noise \( \boldsymbol{w}_k \), while the DVL beams \( \boldsymbol{z}_k^{\mathrm{DVL}} \) are corrupted by measurement noise \( \boldsymbol{v}_k \). Let us denote:
\begin{equation}
\boldsymbol{f}_k = \boldsymbol{f}_k^{\mathrm{true}} + \boldsymbol{w}_k^{f}
\end{equation}
\begin{equation}
\boldsymbol{\omega}_k = \boldsymbol{\omega}_k^{\mathrm{true}} + \boldsymbol{w}_k^{\omega}
\end{equation}
\begin{equation}
\boldsymbol{z}_k^{\mathrm{DVL}} = \boldsymbol{z}_k^{\mathrm{true}} + \boldsymbol{v}_k
\end{equation}
where \( \boldsymbol{w}_k^f \) and \( \boldsymbol{w}_k^\omega \) represent the accelerometer and gyroscopes process noise, respectively, and \( \boldsymbol{v}_k \) denotes the DVL measurement noise.
Due to the nonlinear mapping, \( \mathcal{F}_\theta(\cdot) \), the output velocity estimate becomes a complex function of all the noise sources:
\begin{equation}
\boldsymbol{\hat{v}}_k = \mathcal{F}_\theta\left(\boldsymbol{f}_k^{\mathrm{true}} + \boldsymbol{w}_k^{f},\; \boldsymbol{\omega}_k^{\mathrm{true}} + \boldsymbol{w}_k^{\omega},\; \boldsymbol{z}_k^{\mathrm{true}} + \boldsymbol{v}_k\right).
\end{equation}
Unlike linear estimators, where uncorrelated inputs lead to uncorrelated outputs, the nonlinear dependency structure in \( \mathcal{F}_\theta \) causes interactions between \( \boldsymbol{w}_k \) and \( \boldsymbol{v}_k \). As a result, the effective measurement used for state correction, the output of the DNN, embeds cross-correlations between the process and measurement noise. That is,
\begin{equation}
\mathbb{E}[\boldsymbol{w}_k \boldsymbol{v}_k^\top] \neq \boldsymbol{0},
\end{equation}
\begin{figure*}[h!]
    \centering
    \includegraphics[width=0.8\linewidth]{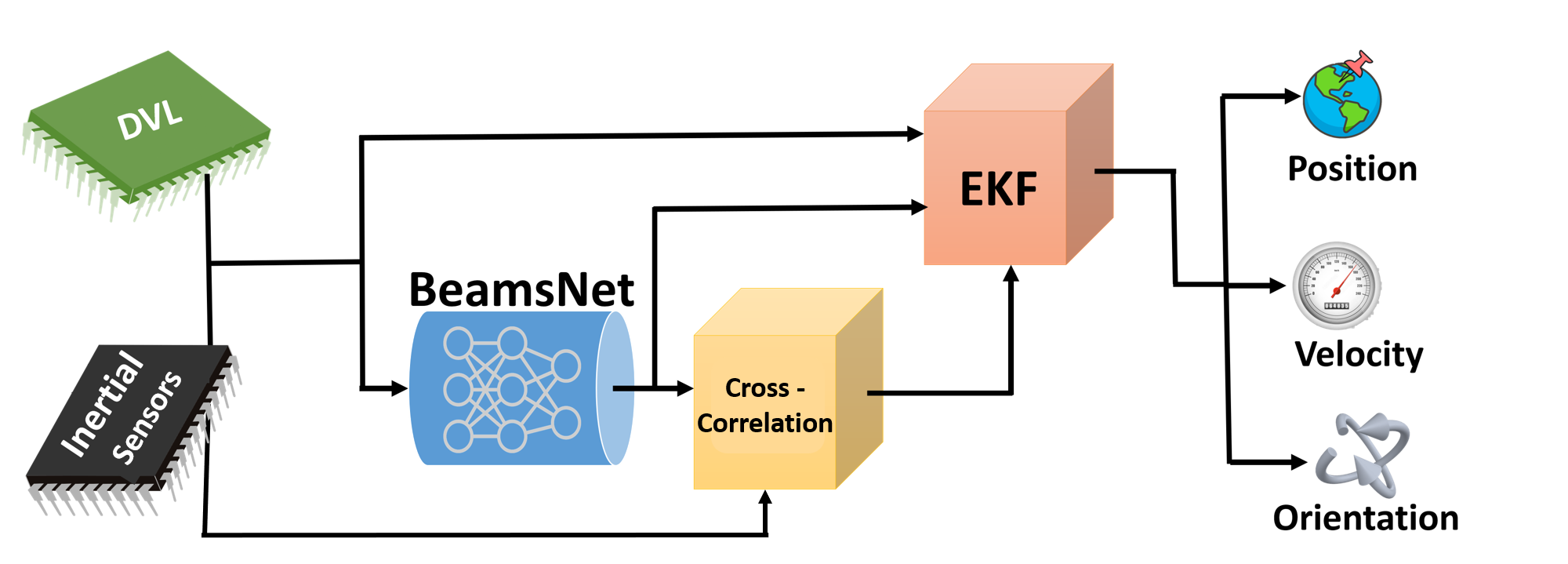}
    \caption{Our proposed approach block diagram illustrates how inertial and DVL measurements are utilized as inputs to the BeamsNet framework. This model generates an enhanced velocity vector measurement update, which is then passed, together with the inertial uncertainty, to the cross-correlation block.  This block computes the cross-correlation matrix, which is subsequently integrated into the EKF to derive the navigation solution.}
    \label{fig:BeamsNet}
\end{figure*}
and the distribution of the estimation error becomes analytically intractable due to the black-box nature of the network. Therefore, when such a fused estimate is used as an update measurement in an error-state EKF, the assumption of noise independence no longer holds. This violates the foundational assumptions of standard Kalman filtering theory and necessitates either reformulation of the filter to incorporate the cross-covariance or empirical techniques to mitigate its effect.
\noindent
\section{Proposed Approach}\label{CS}
\begin{figure}[h!]
    \centering
    \includegraphics[width=\linewidth]{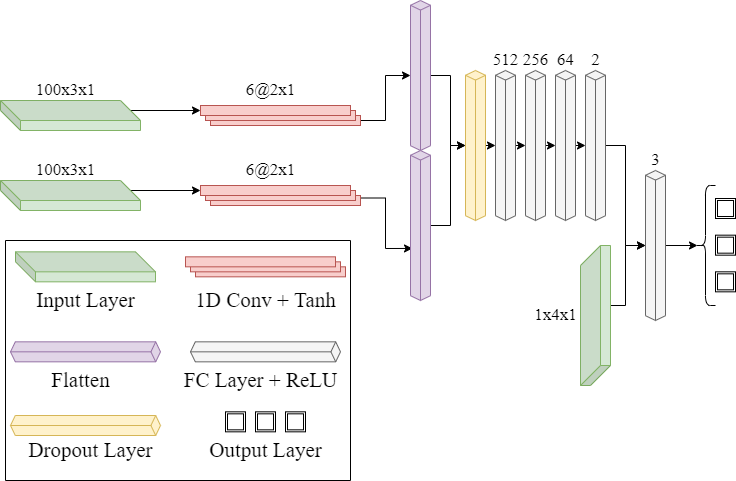}
    \caption{BeamsNet architecture where raw accelerometer and gyroscope measurements are first passed through parallel one-dimensional convolutional layers, each with six filters of size \(2 \times 1\), to extract temporal features from the inertial data. The resulting features are flattened and concatenated, then passed through a dropout layer. A sequence of fully connected layers processes the combined features. Finally, the current DVL measurement is concatenated with the network output and fed into the last fully connected layer, which outputs a \(3 \times 1\) estimated DVL velocity vector.}
    \label{fig:arch}
\end{figure}

\noindent
In this work, we build upon our recent study presented in \cite{cohen2022beamsnet}, where a deep learning-based method, named BeamsNet, was proposed as a replacement for the traditional least squares estimation performed by the DVL, with the goal of providing more accurate velocity measurements. BeamsNet employs a one-dimensional convolutional neural network with a multi-head input architecture that incorporates both past and current DVL beam measurements, as well as inertial data, including the specific force and angular velocity measurements. In this work, we leverage BeamsNet and, together with the inertial reading uncertainty, construct the cross-correlation matrix, which in turn is used in the EKF during the INS/DVL fusion.
Fig.\ref{fig:BeamsNet} presents the information flow within our proposed approach. The BeamsNet architecture and corresponding hyperparameters are presented in Fig.\ref{fig:arch}.

\noindent
As discussed in Section \ref{sec:cross}, fusing the two sensors that traditionally govern the process and measurement models of the error-state EKF within a deep learning framework introduces cross-correlation between the updated velocity measurements produced by BeamsNet and the process noise derived from the inertial sensors. This cross-correlation may degrade the filter’s performance, reduce confidence in the state estimates, and ultimately impact the overall navigation solution. To address this issue, we examine whether accounting for the cross-correlation using the formulation presented in Section \ref{sec:EKFcorr} can improve underwater navigation performance. This approach results in a cross-correlation-aware deep INS/DVL fusion framework. The matrix 
 $\mathbf{M}_k$, defined in \eqref{eqn:M}, plays a key role in this formulation. While the inertial sensor process and measurement noise characteristics are provided in their respective datasheets, BeamsNet uncertainty characteristics are not provided by the network. Therefore, $\mathbf{M}_k$ was determined numerically. To construct a numerical approximation of the cross-covariance between the process noise $\boldsymbol{w}_k \sim \mathcal{N}(\mathbf{0}, \mathbf{Q}_k)$ and the measurement noise $\boldsymbol{v}_k \sim \mathcal{N}(\mathbf{0}, \mathbf{R}_k)$, we assume that both covariance matrices are diagonal and define the cross-correlation matrix as:
\begin{equation}
\mathbb{E}[\boldsymbol{w}_k \boldsymbol{v}_j^\top] = \rho \cdot \sqrt{\operatorname{diag}(\mathbf{Q}_k)} \cdot \left( \sqrt{\operatorname{diag}(\mathbf{R}_k)} \right)^\top 
\label{eq:numerical_cross_cov}
\end{equation}
where $\rho \in [0, 1]$ is a scalar correlation coefficient.
\begin{figure*}[h!]
    \centering

    \begin{subfigure}[b]{0.85\columnwidth}
        \centering
        \includegraphics[width=\linewidth]{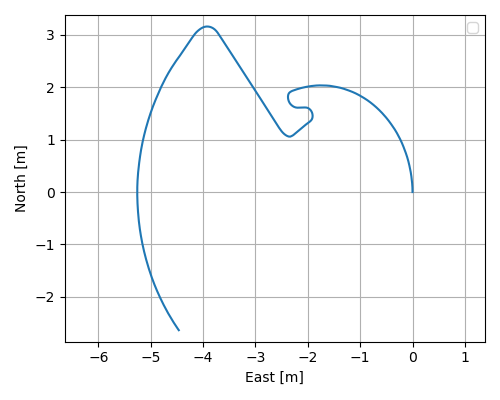}
        \caption{}
        \label{fig:subfig1}
    \end{subfigure}
    \begin{subfigure}[b]{0.85\columnwidth}
        \centering
        \includegraphics[width=\linewidth]{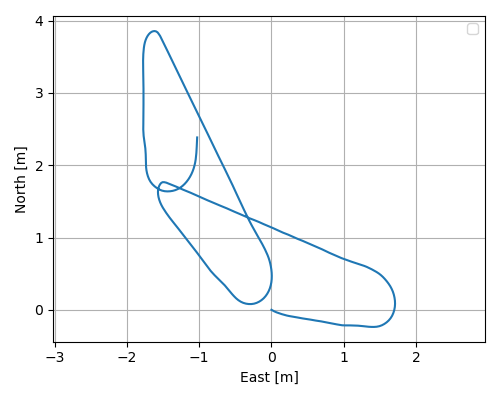}
        \caption{}
        \label{fig:subfig2}
    \end{subfigure}
    \caption{Two AUV trajectories presented in the North-East plane: the trajectory shown in (a) is referred to as "Trajectory \#1," and the trajectory in (b) as "Trajectory \#2." }
    \label{fig:trajs}
\end{figure*}
\section{Analysis and Results}\label{res}
\noindent
To evaluate the performance of the cross-correlation-aware deep INS/DVL fusion, we used an AUV dataset introduced in the original BeamsNet paper, which is also publicly available through the \href{https://github.com/ansfl/BeamsNet}{BeamsNet GitHub repository}.
Approximately four hours of data were collected from an AUV performing various missions in the Mediterranean Sea and used to train and validate the network. This dataset was used to train the BeamsNet network. 
For testing, we utilized a different publicly available dataset introduced in \cite{cohen2025adaptive}, which can be accessed through the corresponding \href{https://github.com/ansfl/A-KIT}{GitHub repository}. We used two distinct 400-second-long missions, each exhibiting different characteristics and sea state conditions. Both were used to first evaluate the robustness of the BeamsNet approach and then assess the performance of the cross-correlation-aware method in comparison to the approach that neglects cross-correlations. The two trajectories used in the evaluation are shown in Fig.\ref{fig:trajs}.
\\ \noindent
To evaluate BeamsNet's performance on the unseen data, we employed the following metrics:
\begin{equation}\label{eqn:6}
    \centering
    \text{RMSE}(\boldsymbol{x}_{\dot\imath},\hat{\boldsymbol{x}}_{\dot\imath})=\sqrt{\frac{\sum_{\dot\imath=1}^{N}(\boldsymbol{x}_{\dot\imath}-\hat{\boldsymbol{x}}_{\dot\imath})^{2}}{N}}
\end{equation}
\begin{equation}\label{eqn:7}
    \centering
        \text{MAE}(\boldsymbol{x}_{\dot\imath},\hat{\boldsymbol{x}}_{\dot\imath})=\frac{\sum_{\dot\imath=1}^{N}|\boldsymbol{x}_{\dot\imath}-\hat{\boldsymbol{x}}_{\dot\imath}|}{N}
\end{equation}
\begin{equation}\label{eqn:8}
    \centering
        \text{R}^{2}(\boldsymbol{x}_{\dot\imath},\hat{\boldsymbol{x}}_{\dot\imath})=1- \frac{\sum_{\dot\imath=1}^{N}(\boldsymbol{x}_{\dot\imath}-\hat{\boldsymbol{x}}_{\dot\imath})^{2}}{\sum_{\dot\imath=1}^{N}(\boldsymbol{x}_{\dot\imath}-\bar{\boldsymbol{x}}_{\dot\imath})^{2}}
\end{equation}
\begin{equation}\label{eqn:9}
    \centering
        \text{VAF}(\boldsymbol{x}_{\dot\imath},\hat{\boldsymbol{x}}_{\dot\imath})=[1-\frac{var(\boldsymbol{x}_{\dot\imath}-\hat{\boldsymbol{x}}_{\dot\imath})}{var(\boldsymbol{x}_{\dot\imath})}]\times100
\end{equation}
in this formulation, \( N \) denotes the total number of samples. The term \( \boldsymbol{x}_{\dot\imath} \) corresponds to the ground truth norm of the DVL-derived velocity vector, while \( \hat{\boldsymbol{x}}_{\dot\imath} \) refers to the predicted velocity norm. The quantity \( \bar{\boldsymbol{x}}_{\dot\imath} \) indicates the average value of the ground truth velocity norm. The function \( \operatorname{var} \) denotes the variance. An ideal model would yield a VAF of 100, an \( R^2 \) value of 1, and both RMSE and MAE equal to zero, reflecting perfect predictive performance. \\
\noindent
The results of the velocity vector estimation are summarized in Table \ref{tab:metrics1} for trajectory \#1 and Table \ref{tab:metrics2} for trajectory \#2.
\begin{table}[h!]
\centering
\caption{Comparison of velocity estimation performance for Trajectory \#1 using the least squares method and the BeamsNet approach. The metrics include RMSE, MAE, coefficient of determination \( R^2 \), and variance accounted for (VAF).}

\label{tab:metrics1}
\resizebox{\columnwidth}{!}{%
\begin{tabular}{|c|c|c|c|c|}
\hline
Method / Metric & RMSE {[}m/s{]} & MAE {[}m/s{]} & $R^2$    & VAF {[}\%{]} \\ \hline
LS              & 0.013494       & 0.012888      & 0.997273 & 99.976048    \\ \hline
BeamsNet (ours) & 0.004015       & 0.003171      & 0.999759 & 99.975878    \\ \hline
\end{tabular}%
}
\end{table}
\begin{table}[h!]
\centering
\caption{Comparison of velocity estimation performance for Trajectory \#2 using the least squares method and the BeamsNet approach. The metrics include RMSE, MAE, coefficient of determination \( R^2 \), and variance accounted for (VAF).}
\label{tab:metrics2}
\resizebox{\columnwidth}{!}{%
\begin{tabular}{|c|c|c|c|c|}
\hline
Method / Metric & RMSE [m/s] & MAE [m/s] & $R^2$    & VAF [\%] \\ \hline
LS              & 0.015122               & 0.014651              & 0.988125 & 99.927172            \\ \hline
BeamsNet (ours) & 0.003927               & 0.003047              & 0.999199 & 99.919904            \\ \hline
\end{tabular}%
}
\end{table}

\noindent
The BeamsNet method maintained high accuracy, which was consistent with the results reported in the original study. The statistical fit of the model, as indicated by the \( R^2 \) and VAF metrics, remained high. In terms of RMSE, BeamsNet achieved an improvement of approximately 70\% over the model-based approach for Trajectory~\#1, and around 74\% for Trajectory~\#2. These results demonstrate the robustness of BeamsNet, as it performs reliably even on previously unseen data.
\begin{figure}[h!]
    \centering
    \begin{subfigure}[b]{0.45\columnwidth}
        \centering
        \includegraphics[width=\linewidth]{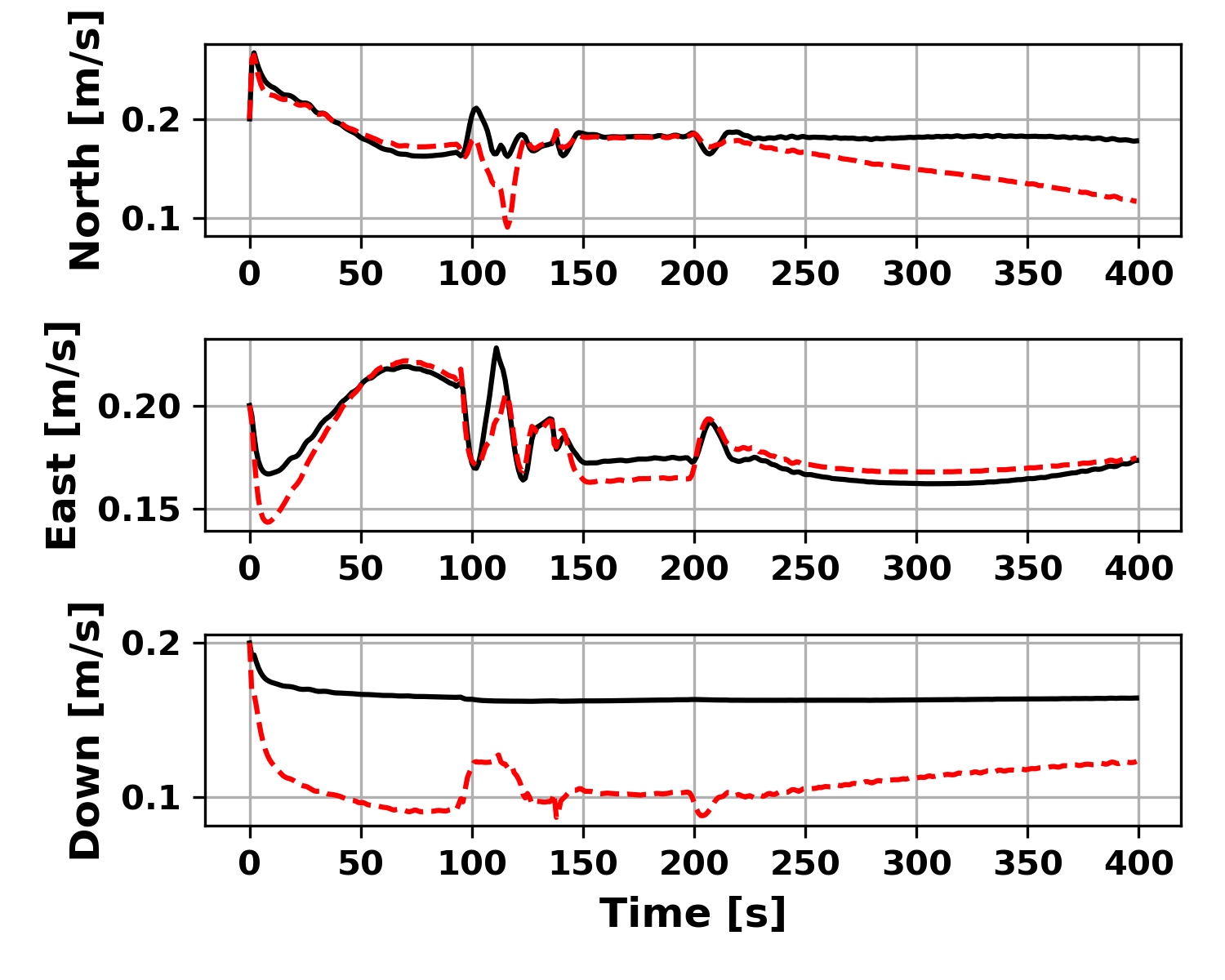}
        \caption{Velocity standard deviation in the North, East, and Down axes.}
        \label{fig:a}
    \end{subfigure}
    \begin{subfigure}[b]{0.45\columnwidth}
        \centering
        \includegraphics[width=\linewidth]{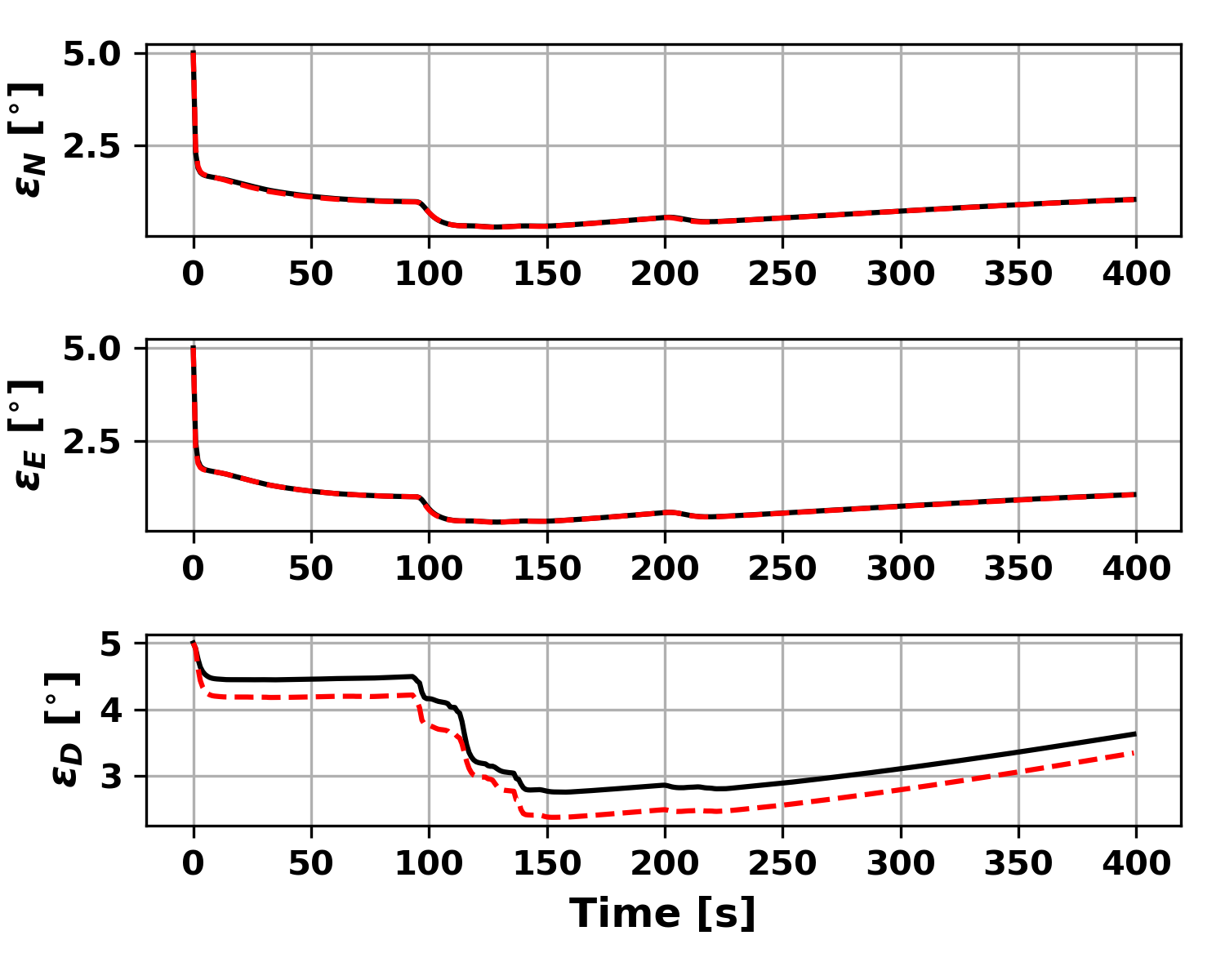}
        \caption{Standard deviation of misalignment angles: roll, pitch, and yaw (from top to bottom).}
        \label{fig:b}
    \end{subfigure}
    
    \vspace{0.5em}
    
    \begin{subfigure}[b]{0.45\columnwidth}
        \centering
        \includegraphics[width=\linewidth]{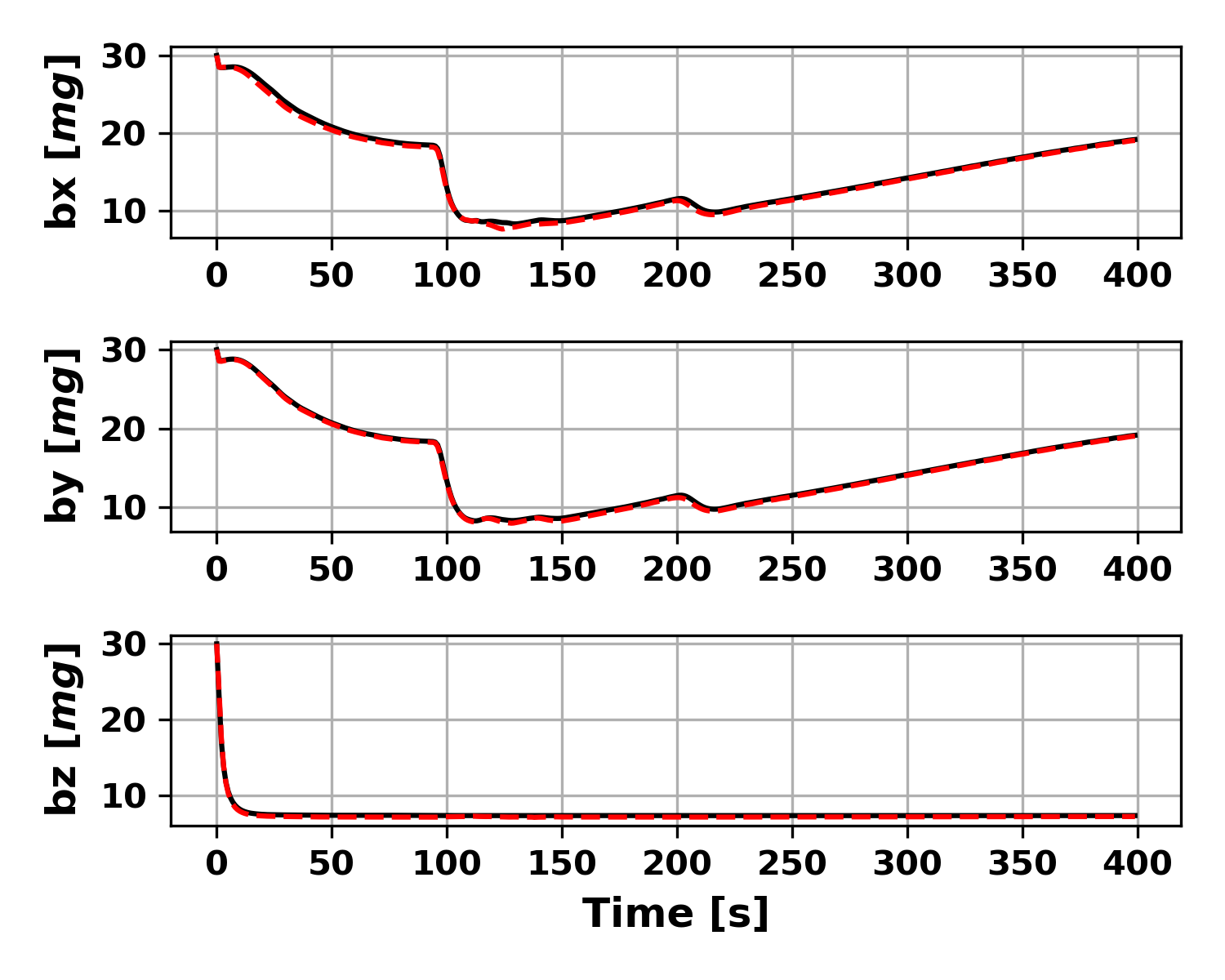}
        \caption{Standard deviation of accelerometer bias estimates in the body frame.}
        \label{fig:c}
    \end{subfigure}
    \begin{subfigure}[b]{0.45\columnwidth}
        \centering
        \includegraphics[width=\linewidth]{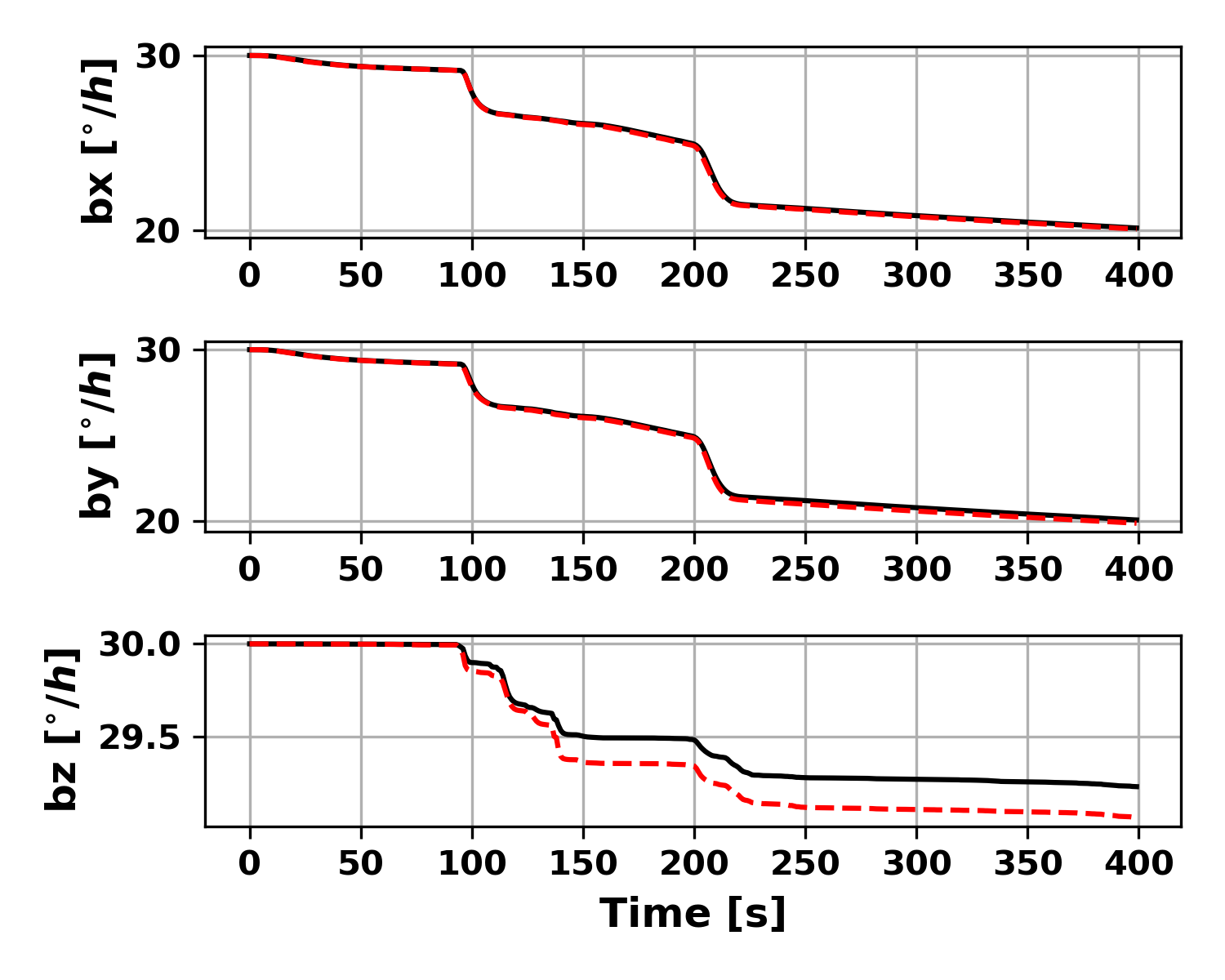}
        \caption{Standard deviation of gyroscope bias estimates in the body frame.}
        \label{fig:d}
    \end{subfigure}
    
    \caption{Trajectory \#1 standard deviation of the states when comparing the cross-correlation-aware approach (red) to the one that neglects it (black).}
    \label{fig:stdTraj1}
\end{figure}
\begin{figure}[h!]
    \centering
    \begin{subfigure}[h]{0.45\columnwidth}
        \centering
        \includegraphics[width = \columnwidth]{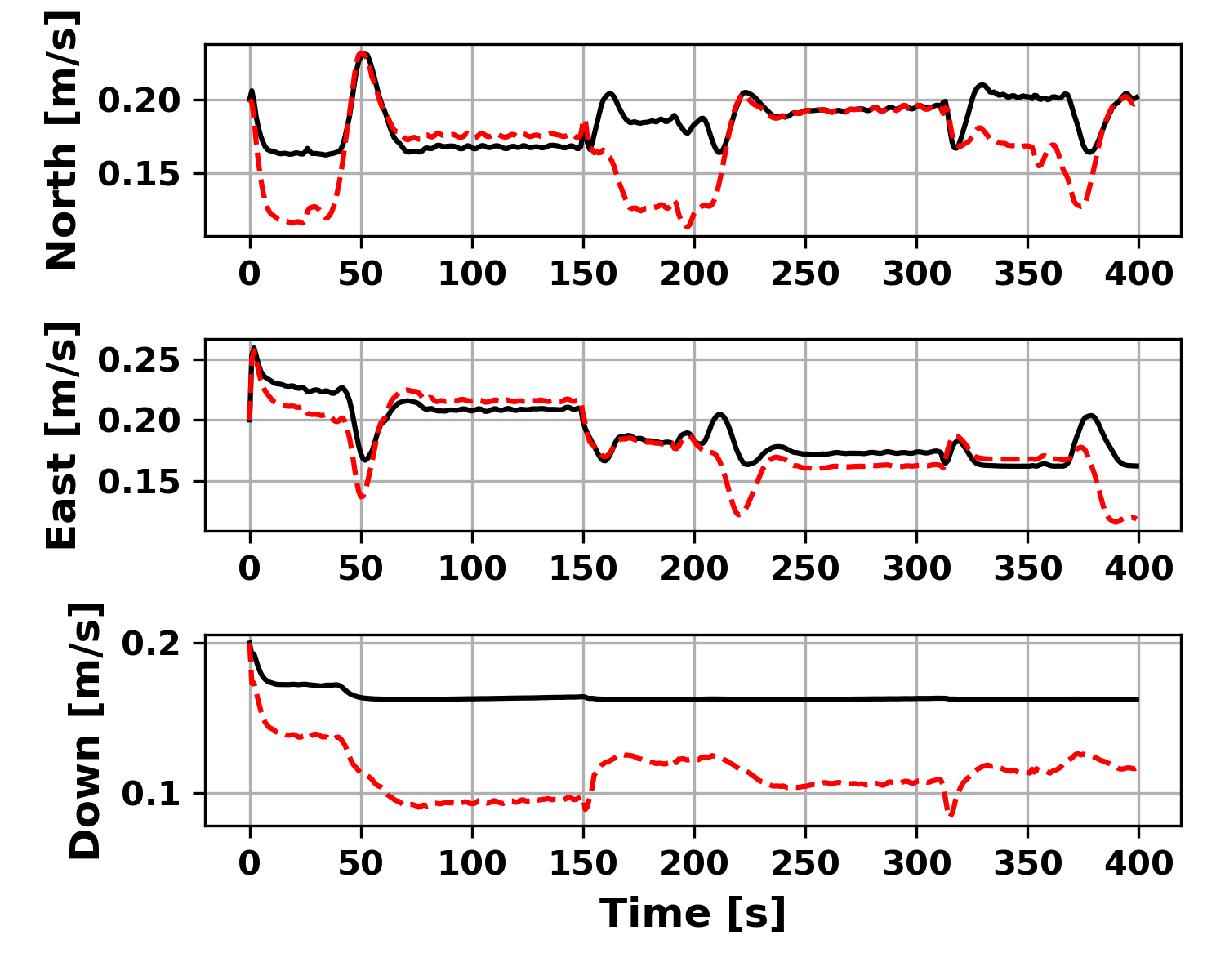}
        \caption{Velocity standard deviation in the North, East, and Down axes.}
        \label{fig:a}
    \end{subfigure}
    \begin{subfigure}[h]{0.45\columnwidth}
        \centering
        \includegraphics[width =\columnwidth]{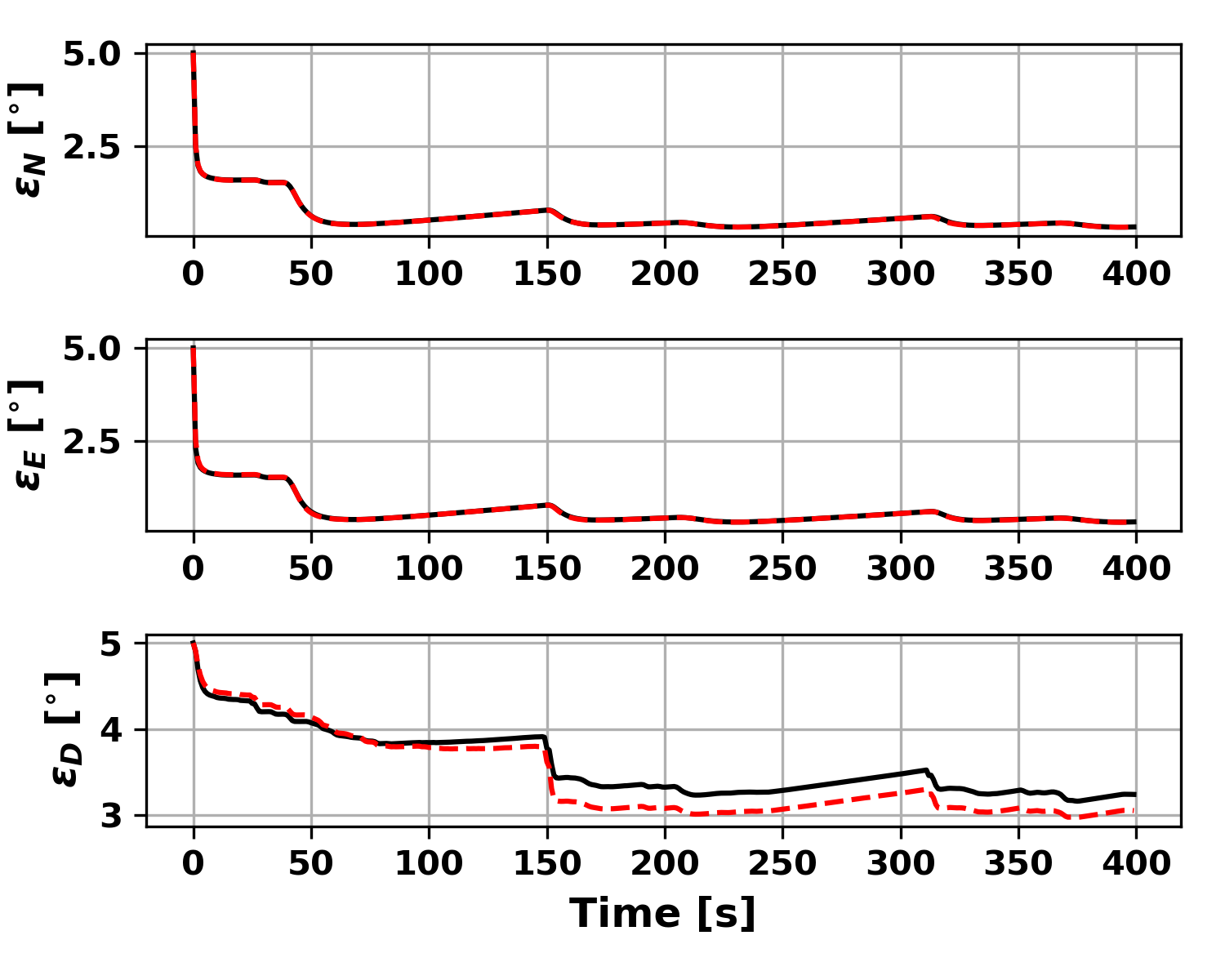}
        \caption{Standard deviation of misalignment angles: roll, pitch, and yaw (from top to bottom).}
        \label{fig:b}
    \end{subfigure}
    
    \vspace{0.5em}
    
    \begin{subfigure}[h]{0.45\columnwidth}
        \centering
        \includegraphics[width=\columnwidth]{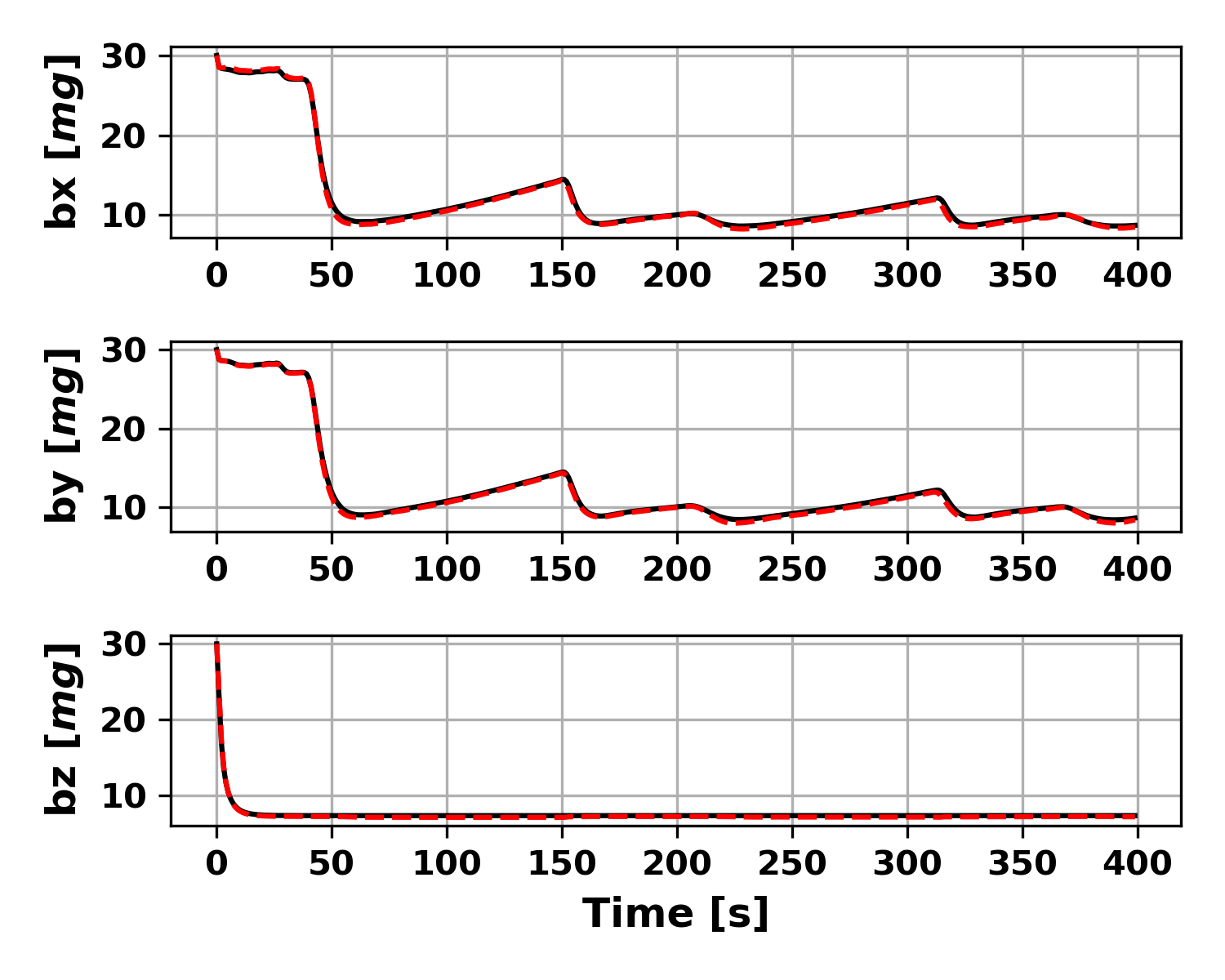}
        \caption{Standard deviation of accelerometer bias estimates in the body frame.}
        \label{fig:c}
    \end{subfigure}
    \begin{subfigure}[h]{0.45\columnwidth}
        \centering
        \includegraphics[width=\columnwidth]{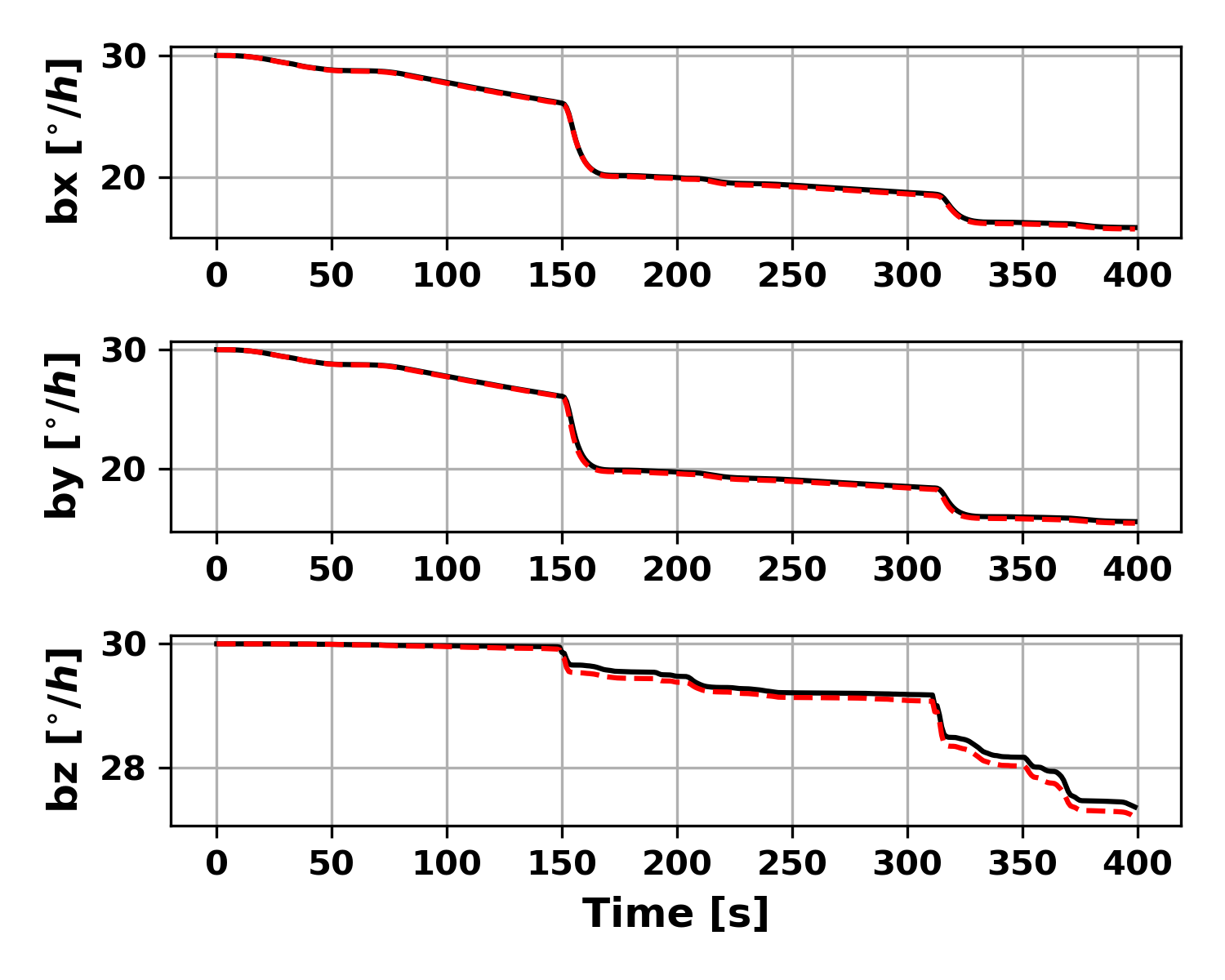}
        \caption{Standard deviation of gyroscope bias estimates in the body frame.}
        \label{fig:d}
    \end{subfigure}
    
    \caption{Trajectory \#2 standard deviation of the states when comparing the cross-correlation-aware approach (red) to the one that neglects it (black).}
    \label{fig:stdTraj2}
\end{figure}

 \noindent
Once the superior performance of the data-driven approach over the least squares method was established, we proceeded to examine the accuracy of the overall navigation solution within the error-state EKF, considering both the case where cross-correlation is neglected and the case where a cross-correlation-aware formulation is employed.
\begin{figure}[b!]
    \centering

    \begin{subfigure}[h]{\columnwidth}
        \centering
        \includegraphics[width=0.95\columnwidth]{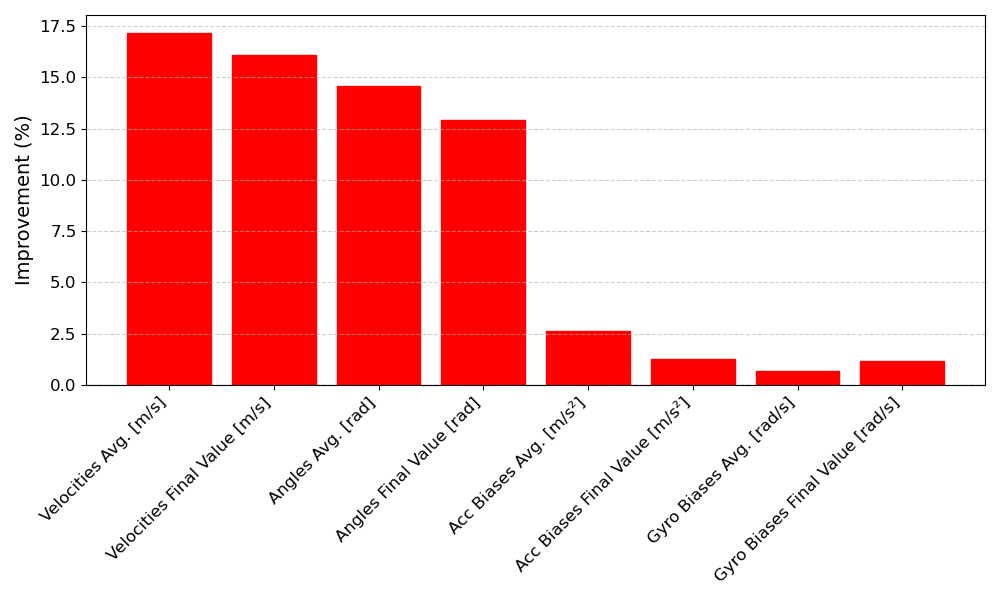}
        \caption{}
        \label{fig:subfig1}
    \end{subfigure}

    \begin{subfigure}[h]{\columnwidth}
        \centering
        \includegraphics[width=0.95\columnwidth]{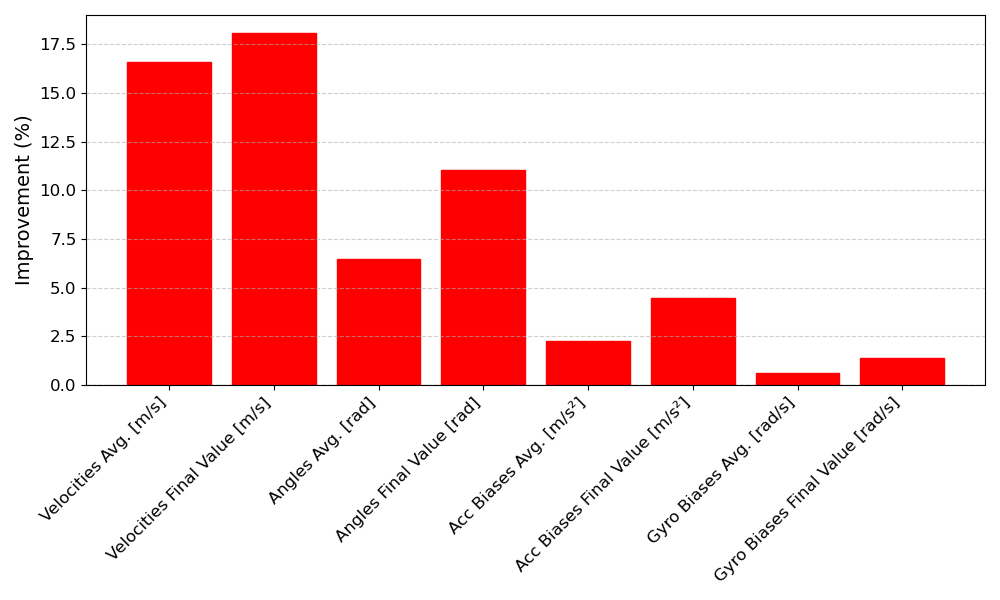}
        \caption{}
        \label{fig:subfig2}
    \end{subfigure}

    \caption{Quantitative evaluation of filter uncertainty for (a) Trajectory~\#1 and (b) Trajectory~\#2. For each three-axis state group (velocity, misalignment angles, accelerometer bias, and gyroscope bias), we compute the average standard deviation over time, as well as the final summed standard deviation across axes. The bars indicate the percentage improvement achieved by the cross-correlation-aware method.}

    \label{fig:bars}
\end{figure}
To begin, we first found the scalar correlation coefficient, which was manually determined through trial and error and set to $\rho = 0.42$. Next, we examined the error in the state vector when comparing with and without the cross-correlation matrix. Those differences were negligible when examining the average error produced by the filters in the two scenarios. However, a notable distinction emerged in the estimated covariance: the cross-correlation-aware method exhibited higher confidence in the state estimation. This resulted in lower values for the error-state covariance.  Fig. \ref{fig:stdTraj1} and Fig. \ref{fig:stdTraj2} present the standard deviations of the velocity, misalignment angles, accelerometer, and gyroscope biases error states for trajectory \#1 and trajectory \#2, respectively. It can be observed that when cross-correlation is properly accounted for, the standard deviation across all states is either comparable or, more frequently, lower in magnitude. This indicates increased confidence in the model's estimates. To quantify this, we examined each group of three-axis states: velocity, misalignment angles, accelerometer bias, and gyroscope bias. For each group, we first computed the average standard deviation of the state estimates over time, across the three axes of each state. Then, we summed the final standard deviation values derived from the estimated covariance matrix across the three axes of each state. These two measures indicate, respectively, whether there is an overall reduction in uncertainty over time and whether the filter converged to a lower value, as well as the percentage of improvement. All findings are summarized in Fig.~\ref{fig:bars}. It can be observed that, for both trajectories, the cross-correlation-aware method outperforms the approach in which cross-correlation is neglected. Notable improvements are observed in the velocity and misalignment angle states, with a general improvement over time exceeding 10\% in most scenarios.

\section{Conclusions}\label{con}
\noindent
This work introduced a cross-correlation-aware deep INS/DVL fusion framework that integrates the strengths of both data-driven and model-based approaches. First, we built upon a previous work called BeamsNet and showed its robustness to unseen data. Then, by incorporating deep learning-based velocity estimates into an error-state EKF with an explicit cross-covariance model, we achieved a solution that is not only superior in terms of accuracy when compared to the model-based least squares approach but also more consistent and theoretically grounded. The proposed method addresses a critical limitation of traditional EKF formulations, namely the assumption of uncorrelated process and measurement noise, a condition often violated when using data-driven measurements. Our results demonstrate that accounting for these correlations yields improved confidence in state estimates and reduced uncertainty over time.
\\ \noindent
Beyond its empirical advantages, this approach offers a principled pathway for integrating modern deep learning techniques within the well-established Kalman filtering framework. This synergy is especially crucial in real-time underwater navigation applications, where reliability, robustness, and theoretical soundness are essential for operational success.
\section*{Acknowledgments}
\noindent
N.C. is supported by the Maurice Hatter Foundation and
University of Haifa presidential scholarship for outstanding students on a direct Ph.D. track.
\bibliographystyle{ieeetr}
\bibliography{refs}

\end{document}